\documentclass[letterpaper,10pt,conference]{ieeeconf}
\IEEEoverridecommandlockouts 

\usepackage{amssymb,amsmath,amscd}
\usepackage{graphicx}
\usepackage[noend]{algorithmic}
\usepackage{xspace}
\usepackage{comment,color}
\usepackage{subfigure}

\newtheorem{theorem}{Theorem}[section]
\newtheorem{defn}[theorem]{Definition}

\newtheorem{lemma}[theorem]{Lemma}

\newtheorem{corollary}[theorem]{Corollary}
\newtheorem{remark}[theorem]{Remark}

\newcommand{\R}{\mathbb{R}}

\newcommand{\N}{\mathbb{N}}
\newcommand{\modulo}[1]{\;(\mathrm{mod}\;{#1})}

\newcommand{\tar}{\mathbf{q}}
\newcommand{\env}{\mathcal{E}}
\newcommand{\p}{\mathbf{p}}
\newcommand{\q}{\mathbf{q}}
\newcommand{\tarset}{\mathcal{Q}}

\newcommand{\I}{\mathcal{I}}
\newcommand{\J}{\mathcal{J}}
\newcommand{\K}{\mathcal{K}}
\newcommand{\tour}{\operatorname{tour}}
\newcommand{\prev}{\textup{prev}}
\newcommand{\cu}{\textup{curr}}
\newcommand{\n}{\textup{next}}
\newcommand{\status}{\textup{status}}
\newcommand{\mess}{\textup{msg}}
\newcommand{\dist}{\textup{dist}}

\newcommand{\etspa}{\textsc{ETSP Assignment}\xspace}
\newcommand{\commrd}{\textsc{comm-rd}\xspace}
\newcommand{\algclass}{assignment-based motion\xspace}
\newcommand{\algclassemph}{\emph{assignment-based motion}\xspace}

\newcommand{\ind}[2]{ {#1}^{[#2]}}
\newcommand{\tind}[2]{ {#1}_{#2}}

\newcommand\oprocendsymbol{\hbox{$\bullet$}}
\newcommand\oprocend{\relax\ifmmode\else\unskip\hfill\fi\oprocendsymbol}

\renewcommand{\theenumi}{(\roman{enumi})}

\begin{document}

\title{Target assignment for robotic networks:\\
  asymptotic performance under limited communication
  \thanks{This material is based upon
    work supported in part by ARO MURI Award W911NF-05-1-0219 and NSF
    SENSORS Award IIS-0330008.}  } \author{Stephen L. Smith \quad%
  Francesco Bullo \thanks{Stephen L. Smith and Francesco Bullo are
    with the Department of Mechanical Engineering, and the Center for
    Control, Dynamical Systems and Computation, University of
    California, Santa Barbara, CA 93106, USA,
    \texttt{\{stephen,bullo\}@engineering.ucsb.edu}}}
\maketitle
\begin{abstract}
We are given an equal number of mobile robotic agents, and distinct target
locations.  Each agent has simple integrator dynamics, a limited
communication range, and knowledge of the position of every target.  We
address the problem of designing a distributed algorithm that allows the
group of agents to divide the targets among themselves and,
simultaneously, leads each agent to reach its unique target.  We do not
require connectivity of the communication graph at any time. We introduce
a novel assignment-based algorithm with the following features: initial
assignments and robot motions follow a greedy rule, and distributed
refinements of the assignment exploit an implicit circular ordering of the
targets. We prove correctness of the algorithm, and give worst-case
asymptotic bounds on the time to complete the assignment as the
environment grows with the number of agents.  We show that among a certain
class of distributed algorithms, our algorithm is asymptotically optimal.
The analysis utilizes results on the Euclidean traveling salesperson
problem.
\end{abstract}

\section{Introduction}
\label{sec:Intro}

\PARstart{C}{onsider} a group of $n$ mobile robotic agents and $n$ target
locations, all lying in $\R^d$, $d \geq 1$.  Each agent has a limited
communication range, and knows the location of some subset (possibly all)
of the $n$ targets through GPS coordinates or a map of the environment.
The \emph{target assignment problem} we consider is to design a
distributed algorithm that allows the group of agents to divide the $n$
targets among themselves and, simultaneously, that leads each agent to
reach its unique target. Such a problem could arise in several
applications.  For example, one could think of the agents as UAV's on a
surveillance mission, and the targets as the centers of their desired
loitering patterns. Or in the context of formation control, the target
positions could describe the desired formation for a group of robots.

The first question is; how do we divide the targets among the agents in a
centralized fashion?  A reasonable strategy would be to minimize the sum
of the distances traveled by each agent to arrive at its target.  The
problem of optimally dividing $n$ persons among $n$ objects, subject to a
linear cost function, is a problem in combinatorial optimization
\cite{BK-JV:05}.  It is referred to as the \emph{assignment problem}, or
the \emph{minimum weight perfect
  matching problem in bipartite graphs}.  The assignment problem can
be written as an integer linear program.  Unlike some integer linear
programs, such as the Euclidean traveling salesperson problem (ETSP),
optimal solutions for the assignment problem can be computed in polynomial
time.  In 1955 Kuhn \cite{HWK:55} developed the Hungarian method---the
first polynomial solution for the assignment problem. Kuhn's method solves
the problem in $O(n^3)$ computation time (see Section \ref{sec:background}
for a definition of the $O$ notation).

Another approach to the assignment problem is the \emph{auction
  algorithm} \cite{DPB-DAC:91,DPB-JNT:97,DAC-CW:03}, first proposed by
Bertsekas.  This method solves the problem in $O(n^3)$ computation time,
but can be computed in a parallel fashion, with one processor for each
person. Recently, Moore and Passino \cite{BJM-KMP:06} modified the auction
algorithm to assign mobile robots to spatially distributed tasks in the
presence of communication delays.  However, in order to exchange bids on a
particular object (task), the auction algorithm, and thus the work in
\cite{BJM-KMP:06}, requires that the communication graph between
processors (robots) is complete.

In this paper we address the target assignment problem when each agent has
knowledge of all target positions, and a limited communication range $r
>0$.  We introduce a class of distributed algorithms, called
\algclassemph, which provide a natural approach to the problem. Following
the recent interest in determining the time complexity of distributed
algorithms for robotic networks, as in \cite{VS-MS-EF-PV:05a} and
\cite{SM-FB-JC-EF:05mn-tmp}, we study the worst-case asymptotic
performance of the \algclass class as the environment grows with $n$.  We
show that for a $d$-dimensional cube environment, $[0,\ell(n)]^d$, $d\geq
1$, if the side length $\ell(n)$ grows at a rate of at least
$(1+\epsilon)rn^{1/d}$, where $\epsilon
>0$, then the completion time is in $\Omega(n^{(d-1)/d}\ell(n))$, for
all algorithms in this class.

In Section \ref{sec:ETSP_ASST} we introduce a novel control and
communication algorithm, called \etspa. In this algorithm, each agent
computes an ETSP tour through the $n$ targets, turning the cloud of target
points into an ordered ring. Agents then move along the ring, looking for
the next available target.  When agents communicate, they exchange
messages of $O(\log{n})$ size, containing information on how far it is to
the next available target along the ring.  In Section
\ref{sec:correctness}, we verify the correctness of this algorithm for any
communication graph which contains, as a subgraph, the $r$-disk graph.  In
Section \ref{sec:time_complex}, we show that when $\ell(n)\geq
(1+\epsilon)rn^{1/d}$, among all algorithms in the \algclass class, the
\etspa algorithm is asymptotically optimal (i.e., a constant factor
approximation of the optimal).  Finally, in Section \ref{sec:n_neq_m}, we
note that \etspa solves the target assignment problem even when there are
$n$ agents and $m$ targets, $n\neq m$.

\section{Background}
\label{sec:background} In this section we introduce notation and review
some relevant results in combinatorial optimization.
\subsection{Notation}
\label{sec:notation} We let $\R$ denote the set of real numbers, $\R_{>0}$
denote the set of positive real numbers, and $\N$ denote the set of
positive integers.  For a set finite $A$ we let $|A|$ denote its
cardinality.  For two functions $f,g:\N\to\R_{> 0}$, we write $f(n) \in
O(g)$ (respectively, $f(n)\in\Omega(g)$) if there exist $N\in\N$ and
$c\in\R_{>0}$ such that $f(n) \leq cg(n)$ for all $n \geq N$
(respectively, $f(n) \geq cg(n)$ for all $n \geq N$). If $f(n) \in O(g)$
and $f(n) \in \Omega(g)$ we say $f(n) \in \Theta(g)$.  Finally, we use the
notation $\modulo{n}$ to denote arithmetic performed modulo $n\in\N$.
Thus, for an integer $n\in\N$ we have $n+1=1\modulo{n}$ and
$0=n\modulo{n}$, and $\{n-1,n,n+1\}=\{n-1,n,1\}\modulo{n}$.

\subsection{The assignment problem}
\label{sec:assign_prob} Following \cite{DPB-JNT:97}, the classical
assignment problem can be described as follows.  Consider $n$ persons who
wish to divide themselves among $n$ objects.  For each person $i$, there
is a nonempty set $\ind{\tarset}{i}$ of objects that $i$ can be assigned
to, and cost $c_{ij}\geq 0$ associated to each object
$j\in\ind{\tarset}{i}$.  An \emph{assignment} $S$ is a set of
person-object pairs $(i,j)$ such that $j\in \ind{\tarset}{i}$ for all
$(i,j)\in S$.  For each person $i$ (likewise, object $j$), there is at
most one pair $(i,j)\in S$.  We call the assignment \emph{complete} if it
contains $n$ pairs.  The goal is to find the complete assignment which
minimizes $\sum_{(i,j)\in S}c_{ij}$.

Let $x_{ij}$ be a set of variables for $i$ and $j$ in
$\I:=\{1,\ldots,n\}$.  For an assignment $S$, we write $x_{ij}=1$ if
$(i,j)\in S$, and $x_{ij}=0$ otherwise.  Thus, the problem of determining
the optimal assignment can be written as a linear program:
\begin{align*}
  \text{minimize}& \quad \sum_{i=1}^n\sum_{j\in \ind{\tarset}{i}}c_{ij}x_{ij},\\
  \text{subject to}& \quad \sum_{j\in \ind{\tarset}{i}}x_{ij} =1\quad \forall
  \; i\in\I, \\
  &\quad \sum_{\{i|j\in \ind{\tarset}{i}\}}x_{ij} =1\quad \forall \; j\in\I, \\
  &\quad x_{ij}\geq 0.
\end{align*}
We cannot use linear inequalities to write the constraint that $x_{ij}$'s
attain only the values zero and one.  However, it turns out,
\cite{DPB-JNT:97}, that there always exists an optimal solution in which
the $x_{ij}$'s satisfy our integer assumption.

\subsection{The Euclidean traveling salesperson problem}
\label{sec:TSP_review} Here we review some relevant results on the
Euclidean traveling salesperson problem (ETSP).  Let $\tarset$ be a set of
$n$ points in a compact environment $\env\subset \R^d$, $d\geq 1$, and let
$\tarset_n$ be the set of all point sets $\tarset\subset\env$ with
$|\tarset|=n$. Let $\mathrm{ETSP}(\tarset)$ denote the cost of the ETSP
tour over the point set $\tarset$, i.e., the length of the shortest closed
path through all points in $\tarset$.  An important result, from
\cite{KJS-EMR-DAP:83}, is that given a compact set $\env$, there exists a
finite constant $\alpha(\env)$ such that, for all $\tarset\in \tarset_n$,
\begin{equation}
\label{eq:ETSP_bound} \mathrm{ETSP}(\tarset) \leq \alpha(\env)n^{(d-1)/d}.
\end{equation}
In fact, we have that in the worst-case setting, the
$\mathrm{ETSP}(\tarset)$ belongs to $\Theta(n^{(d-1)/d})$.

In our application of these results it will be useful to consider the case
where the environment grows with the number of points.  That is, we are
interested in environments which are cubes, $[0,\ell(n)]^d$, $d\geq 1$,
where $\ell(n)$ is the side length of the cube. Applying a simple scaling
argument to the result in \eqref{eq:ETSP_bound}, we arrive at the
following corollary.
\begin{corollary}[ETSP tour length]
\label{thm:tsp_grow} Consider an environment $\env=[0,\ell(n)]^d$, where
$d\geq 1$. For every point set $\tarset\in \tarset_n$,
\[
\mathrm{ETSP}(\tarset) \in O(n^{(d-1)/d}\ell(n)).
\]
\end{corollary}

The problem of computing an optimal tour is known to be NP-complete.
However, there exist heuristics which can be computed efficiently and give
a constant factor approximation to the optimal tour.  The best known
approximation algorithm is due to Christofides \cite{NC:76}. The
\emph{Christofides' algorithm} computes a tour that is no more than $3/2$
times longer than the optimal.  It runs in time $O(n^3)$. Another method,
known as the \emph{double-tree algorithm}, produces tours that are no
longer than twice the optimal, in run time $O(n^2)$.

\section{Problem formulation}
\label{sec:prob_form} To describe the target assignment problem formally,
consider $n$ agents in an environment $\env(n)\subset\R^d$, $d\geq 1$.
The environment $\env(n)$ is compact for each $n$ but may grow with the
number of agents.  For ease of presentation let $\env:=[0,\ell(n)]^d$,
where $\ell(n) >0$ (that is, $\env$ is a $d$-dimensional cube with side
length $\ell(n)$).  Each agent has a unique identifier (UID) taken from
the set $I_{UID}\subseteq\N$.  For simplicity, we assume that
$I_{UID}:=\I=\{1,\ldots,n\}$.  However, each agent does not know the set
of UIDs being used (i.e., agent $n$ does not know it has the largest UID).
Agent $i\in\I$ has position $\ind{\p}{i}\in\env$.  Two agents, $i$ and $k$
in $\I$, are able to communicate if and only if
$\|\ind{\p}{i}-\ind{\p}{k}\| \leq r$, where $r > 0$ is called the
communication range.  We refer to the graph representing the communication
links as the $r$-disk graph.  Agent $i$'s kinematic model is $\ind{\dot
\p}{i} = \ind{\mathbf{u}}{i}$, where $\ind{\mathbf{u}}{i}$ is a velocity
control input bounded by $v>0$. We assume that the agents move in
continuous time and communicate according to a discrete time communication
schedule consisting of an increasing sequence of time instants with no
accumulation points, $\{t_k\}_{k\in \N}$.  We assume that $|t_{k+1}-t_k|
\leq t_{max}$, for all $k\in \N$, where $t_{max}\in\R_{>0}$.  At each
communication round, agents can exchange messages of length $O(\log{n})$.
\footnote{The number of bits required to represent an ID, unique among
  $n$ agents, grows with the logarithm of $n$.}  We assume that
communication round $k$ occurs at time $t_{k}$, and that all messages are
sent and received instantaneously at $t_{k}$.  Motion then occurs from
$t_k$ until $t_{k+1}$.  It should be noted that in this setup we are
emphasizing the time complexity due to the motion of the agents.

Let $\tarset:=\{\tind{\q}{1},\ldots,\tind{\q}{n}\}$ be a set of distinct
target locations, $\tind{\q}{j}\in \env$ for each $j\in\I$. Agent $i$ is
equipped with memory $\ind{M}{i}$, of size $|\ind{M}{i}|$.  In this
memory, agent $i$ stores a set of target positions,
$\ind{\tarset}{i}\subseteq\tarset$.  These are the targets to which agent
$i$ can be assigned.  We let $\ind{\tarset}{i}(0)$ denote agent $i$'s
initial target set.  These positions may be known through GPS coordinates,
or through a map of the environment.

In this paper we assume that each agent knows the position of every
target.  That is, $\tarset^{[i]}(0)=\tarset$ for each $i\in\I$.  We refer
to this as the \emph{full knowledge} assumption.  To store this amount of
information we must assume that the size of each agents' memory,
$|\ind{M}{i}|$, grows linearly with $n$.  Our goal is to solve the
\emph{full knowledge target assignment problem}:
\begin{quote}
  Determine a control and communication law for $n\in\N$ agents, with
  attributes as described above, satisfying the following requirement.
  There exists a time $T>0$ such that for every agent $i\in\I$, there
  is a unique target $\tind{\q}{j_i}\in\ind{\tarset}{i}(0)$ with
  $\ind{\p}{i}(t)=\tind{\q}{j_i}$ for all time $t\geq T$, where
  $j_i=j_k$ if and only if $i=k$.
\end{quote}
In the remainder of the paper, we will refer to this as the \emph{target
assignment problem}.
\begin{remark}[Consistent knowledge]
\label{rem:consist_knowledge} A more general assumption on the initial
target sets, $\tarset^{[i]}(0)$, which still ensures the existence of a
complete assignment, is the \emph{consistent knowledge} assumption: For
each $\K\subseteq\I$, $\left|\cup_{k\in \K}\tarset^{[k]}(0)\right| \geq
|\K|$.  In fact, it was proved by Frobenius, 1917, and Hall, 1935 that
this is the necessary and sufficient condition for the existence of a
complete assignment \cite{BK-JV:05}.  \oprocend
\end{remark}
In the full knowledge assumption, each agent knows the position of all
targets in $\tarset$.  These positions will be stored in an array within
each agents memory, rather than as an unordered set.  To represent this,
we replace the target set $\tarset$ with the target $n$-tuple
$\tar:=(\q_1,\ldots,\q_n)$, and the local target set $\ind{\tarset}{i}$
with the $n$-tuple $\ind{\tar}{i}$.  Thus, in the full knowledge
assumption, $\ind{\q}{i}(0):=\tar$ for each $i\in\I$. (It is possible that
the order of the targets in the local sets $\ind{\q}{i}$ may initially be
different.  However, given a set of distinct points in $\R^d$, it is
always possible to create a unique ordering.)

\section{Assignment-based algorithms with lower bound analysis}
\label{sec:algclass}

In this section we introduce and analyze a class of deterministic
algorithms for the target assignment problem.

\subsection{The \algclass class}
The initialization, motion, and communication for each algorithm in the
\algclassemph class have the following attributes:
\paragraph*{Initialization} In this class of algorithms agent $i$
initially selects the closest target in $\ind{\tar}{i}$, and sets the
variable $\ind{\cu}{i}$ (agent $i$'s current target), to the index of that
target.
\paragraph*{Motion}Agent $i$ moves toward the target $\ind{\cu}{i}$ at
speed~$v$:
\begin{equation}
\label{eq:cont_law} \ind{\dot\p}{i} =
\begin{cases}
v\frac{\ind{\tar}{i}_{\ind{\cu}{i}}-\ind{\p}{i}}{\|\ind{\tar}{i}_{\ind{\cu}{i}}
-\ind{\p}{i}\|},& \text{if}\; \ind{\tar}{i}_{\ind{\cu}{i}} \neq \ind{\p}{i}, \\
0,& \text{otherwise},
\end{cases}
\end{equation}
where $v > 0$ is a constant.
\paragraph*{Communication} As agent $i$ communicates with other
agents, it updates the tuple $\ind{\tar}{i}$ ``removing'' targets which
are assigned to other agents.  If agent $i$ must change $\ind{\cu}{i}$, it
selects a new target in $\ind{\tar}{i}$, that has not been removed.  This
is described more formally in the following.

\begin{center}
  \noindent{\framebox[.9999\linewidth]{\noindent\parbox{.95\linewidth-2\fboxsep}{
        \textbf{\footnotesize Communication round for agent $i$.}
        \footnotesize
\begin{algorithmic} [1]
  \STATE Broadcast a message, $\ind{\mess}{i}$, based on
  $\ind{\tar}{i}$ and containing $\ind{\cu}{i}$ and the UID $i$.
  \STATE Receive $\ind{\mess}{k}$ from each agent $k$ within
  communication range.
\FORALL{$\ind{\mess}{k}$ received}
  \STATE Based on $\ind{\mess}{k}$, (possibly) remove assigned targets from
  $\ind{\tar}{i}$.
\IF{$\ind{\cu}{i}=\ind{\cu}{k}$}
  \STATE If agent $i$ is farther from $\ind{\cu}{i}$ than agent $k$,
  or if they are the same distance but $i < k$, remove the target given by
  $\ind{\cu}{i}$ from $\ind{\tar}{i}$.
\ENDIF
\ENDFOR
\STATE Set $\ind{\cu}{i}$ to a target in $\ind{\tar}{i}$ (i.e., a
target that has not been removed).
\end{algorithmic}
}}}
\end{center}

\subsection{Lower bound on task complexity}
\label{sec:lowerbound} In order to classify the time complexity of the
\algclass class of algorithms, we introduce a few useful definitions.  We
say that agent $i\in\I$ is \emph{assigned} to target $\ind{\tar}{i}_j$,
$j\in\I$, when $\ind{\cu}{i}=j$.  In this case, we also say target $j$ is
assigned to agent $i$.  We say that agent $i\in\I$ \emph{enters a
  conflict} over the target $\ind{\cu}{i}$, when agent $i$ receives a
message, $\ind{\mess}{k}$, with $\ind{\cu}{i}=\ind{\cu}{k}$.  Agent $i$
\emph{loses the conflict} if agent $i$ is farther from $\ind{\cu}{i}$ than
agent $k$, and \emph{wins the conflict} if agent $i$ is closer to
$\ind{\cu}{i}$ than agent $k$, where ties are broken by comparing UIDs.

Now we show that if agent $i$ is assigned to the same target as another
agent, it will enter a conflict in finite time.
\begin{lemma}[Conflict in finite time]
\label{lem:com_range} Consider any communication range $r>0$, and any
fixed number of agents $n\in \N$. If, for two agents $i$ and $k$,
$\ind{\cu}{i}=\ind{\cu}{k}$ at some time $t_1\geq 0$, then agent $i$ (and
likewise, agent $k$) will enter a conflict over $\ind{\cu}{i}$ in finite
time.
\end{lemma}
\begin{proof}
  For each $n$ the region $\env$ is compact, and the motion for each
  agent is given by \eqref{eq:cont_law}.  Hence, agent $i$ will reach
  $\ind{\cu}{i}$ in no more than $\mathrm{diam}(\env)/v$ time units,
  as will agent $k$.  The condition $\|\ind{\p}{i}-\ind{\p}{k}\|\leq
  r$ will be satisfied within $\mathrm{diam}(\env(n))/v$ time units.
  At the next communication round, agent $i$ will enter a conflict
  over $\ind{\cu}{i}$.
\end{proof}

With these definitions we give a lower bound on the time complexity of the
task assignment problem when the environment grows with the number of
agents.
\begin{theorem}[Time complexity of target assignment]
\label{thm:comp_lower} Consider $n$ agents, with communication range $r
>0$, in an environment $\env=[0,\ell(n)]^d$, $d\geq 1$.  If
$\ell(n)\geq(1+\epsilon)r n^{1/d}$, where $\epsilon \in\mathbb{R}_{>0}$,
then for all algorithms in the \algclass class, the time complexity of the
target assignment problem is in $\Omega(n^{(d-1)/d}\ell(n))$.
\end{theorem}
\begin{proof}
  We will construct a set of target positions and a set of initial
  agent positions for which the bound holds.  The environment $\env$
  is the $d$-cube, $[0,\ell(n)]^d$.  Divide the cube $\env$ into
  $(\lceil n^{1/d} \rceil)^d$ cubes, each with side length
  $\ell(n)/\lceil n^{1/d} \rceil$, and place a target at the center of
  each of the cubes until you run out.  This is shown in Fig.
  \ref{fig:grid}.
\begin{figure}
\begin{center}
\includegraphics[width=0.75\linewidth]{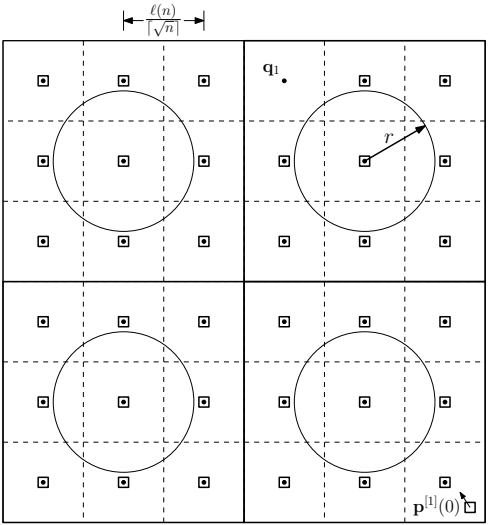}
\caption{Targets and agents placed on a lattice for the proof of
  Theorem \ref{thm:comp_lower}.  The lattice is split into 4 blocks,
  each containing $3^2=9$ agents.  The center agent of each block is
  shown along with its communication radius $r$.  The distance between
  these center agents is lower bounded by $\ell(n)/\lceil n^{1/2}
  \rceil$.}
\label{fig:grid}
\end{center}
\end{figure}
Notice that the distance between any two targets is lower bounded by
$\ell(n)/\lceil n^{1/d} \rceil$, and that, for sufficiently large $n$,
$\ell(n)/\lceil n^{1/d} \rceil \geq (1+\epsilon)rn^{1/d}/\lceil n^{1/d}
\rceil > r$.

Next, place agent $2$ at $\q_2$, agent 3 at $\q_3$ and so on so that
$\ind{\p}{i}=\q_i$, for all $i\in\{2,\ldots,n\}$.  From the
initialization, we have that $\ind{\cu}{i}=i$ for each
$i\in\{2,\ldots,n\}$.  Now, if we place agent 1 in
$\env\setminus\{\tar_1,\ldots,\tar_n\}$, it will lose a conflict over any
of the $n-1$ occupied targets $\tind{\q}{2},\ldots,\tind{\q}{n}$. Thus,
the assignment will not be complete until agent 1 reaches target $\q_1$.
Since the distance between targets is greater than $r$, communication
between agents $i$ and $k$ is not possible for any $i,k\in\{2,\ldots,n\}$.
So, agent $i\in \{2,\ldots,n\}$ will communicate only with agent 1.  Thus,
prior to communication with agent 1, each agent $i\in\{2,\ldots,n\}$ will
have $\ind{\tar}{i}=\tar$. The first time agent $1$ comes within distance
$r$ of a target $j\in\{2,\ldots,n\}$, in the best-case, agent $1$ will
remove target $j$ from $\ind{\tar}{i}$.  Now, for any deterministic method
of selecting $\ind{\cu}{i}$, we can place agent $1$ in
$\env\setminus\{\tar_1,\ldots,\tar_n\}$ such that target $\tar_1$ is the
last target for which agent $1$ will come within distance $r$. Therefore,
agent 1 must come within distance $r$ of each of the $n-1$ assigned
targets, before finally arriving at $\tind{q}{1}$.

Now we will lower bound the distance traveled by agent 1.  To do this,
split the large $d$-cube into $\lfloor n/3^d \rfloor$ smaller $d$-cubes,
or blocks, where each block contains $3^{d}$ targets.  An example is shown
in Fig. \ref{fig:grid}.  There is one target at the center of each of
these blocks, and agent 1 must come within distance $r$ of it.  The
distance between the center target of each block is lower bounded by the
distance between targets, $\ell(n)/\lceil n^{1/d}\rceil$.  Agent $1$ must
travel this distance at least $\lfloor n/3^d \rfloor-1$ times.  So we have
\[
\text{Path length}\geq
\left(\left\lfloor\frac{n}{3^d}\right\rfloor-1\right)\frac{\ell(n)}{\lceil
  n^{1/d}\rceil}\in \Omega(n^{(d-1)/d}\ell(n)).
\]
Hence, the path length is in $\Omega(n^{(d-1)/d}\ell(n))$.  Since
$v\in\R_{>0}$, the time complexity is also in
$\Omega(n^{(d-1)/d}\ell(n))$.
\end{proof}
\begin{remark}[$\ell(n)\leq \ell_{crit}$]
We have lower bounded the time complexity when $\ell(n)$ grows faster than
some critical value, $\ell_{crit}=r n^{1/d}$.  This same type of bound
appears in percolation theory and the study of random geometric graphs,
where it is referred to as the thermodynamic limit \cite{MP:03}.  When
$\ell(n)\leq \ell_{crit}$, congestion issues in both motion and
communication become more prevalent, and a more complex communication and
motion model would ideally be used. \oprocend
\end{remark}
In the next section we introduce an asymptotically optimal algorithm in
the \algclass class.

\section{The \etspa Algorithm}
\label{sec:ETSP_ASST} In this section we introduce the \etspa
algorithm---an algorithm within the \algclass class.  We will show that
when $\ell(n)$ grows more quickly than a critical value, this algorithm is
asymptotically optimal.  The algorithm can be described as follows.

\begin{figure}
\begin{center}
\includegraphics[width=0.80\linewidth]{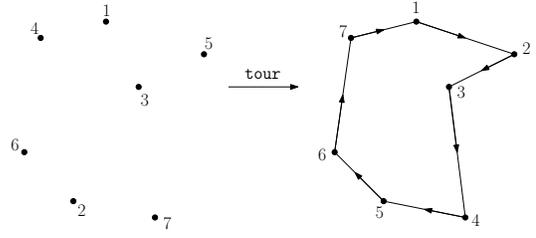}
\caption{The map $\tour$, creating an ETSP tour of seven targets.}
\label{fig:tour}
\end{center}
\end{figure}

\begin{figure}
\begin{center}
\includegraphics[width=0.45\linewidth]{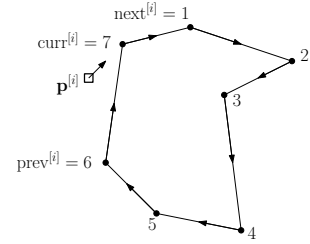}
\caption{The initialization for agent $i$.} \label{fig:tour_i}
\end{center}
\end{figure}

For each $i\in\I$, agent $i$ computes a constant factor approximation of
the optimal ETSP tour of the $n$ targets in $\ind{\tar}{i}$, denoted
$\tour(\ind{\tar}{i})$.  We can think of $\tour$ as a map which reorders
the indices of $\ind{\tar}{i}$; $\tour(\ind{\tar}{i})
=(\ind{\q}{i}_{\sigma(1)},\ldots,\ind{\q}{i}_{\sigma(n)})$, where
$\sigma:\I\to\I$ is a bijection.  Notice that this map is independent of
$i$ since all agents use the same method.  An example is shown in
Fig.~\ref{fig:tour}.
Agent $i$ then replaces its $n$-tuple $\ind{\tar}{i}$ with
$\tour(\ind{\tar}{i})$.  Next, agent $i$ computes the index of the closest
target in $\ind{\tar}{i}$, and calls it $\ind{\cu}{i}$.  Agent $i$ also
maintains the index of the next target in the tour which may be available,
$\ind{\n}{i}$, and first target in the tour before $\ind{\cu}{i}$ which
may be available, $\ind{\prev}{i}$.  Thus, $\ind{\n}{i}$ is initialized to
$\ind{\cu}{i}+1 \modulo{n}$ and $\ind{\prev}{i}$ to $\ind{\cu}{i}-1
\modulo{n}$.  This is depicted in Fig.~\ref{fig:tour_i}.
In order to ``remove'' assigned targets from the tuple $\ind{\tar}{i}$,
agent $i$ also maintains the $n$-tuple, $\ind{\status}{i}$.  Letting
$\ind{\status}{i}(j)$ denote the $j$th entry in the $n$-tuple, the entries
are given by
\begin{equation}
\label{eq:status} \ind{\status}{i}(j) =
\begin{cases}
0,& \begin{aligned} &\text{if agent $i$ knows $\ind{\tar}{i}_j$ is
assigned} \\*[-0.16cm] &\text{to another agent},
\end{aligned} \\
1,& \text{otherwise}.
\end{cases}
\end{equation}
Thus, $\ind{\status}{i}$ is initialized as the $n$-tuple $(1,\ldots,1)$.
The initialization is summarized in Table~\ref{tab:init}.
\begin{table}[tbh]
\caption{\label{tab:init} The initialization procedure for agent $i$.}
\noindent{ \framebox[.9999\linewidth]{ \noindent
\parbox{.99\linewidth-2\fboxsep}{
\textbf{Initialization for agent $i$.}\\*[0.1cm]
        \footnotesize
        \textbf{Assumes:}  $\ind{\tar}{i}:=\tar$ for each $i\in\I$.
\begin{algorithmic}[1]
\STATE Compute a TSP tour of $\ind{\tar}{i}$, $\tour(\ind{\tar}{i})$,
    and set $\ind{\tar}{i}:=\tour(\ind{\tar}{i})$.
\STATE Compute the closest target in $\ind{\tar}{i}$, and set $\ind{\cu}{i}$
    equal to its index:  $\ind{\cu}{i}:= \arg\min_{j\in\I}
        \{\|\ind{\tar}{i}_j-\ind{\p}{i}\|\}$.
\STATE Set $\ind{\n}{i}:=\ind{\cu}{i}+1 \modulo{n}$.
\STATE Set $\ind{\prev}{i}:=\ind{\cu}{i}-1 \modulo{n}$.
\STATE Set $\ind{\status}{i}:=\mathbf{1}_n$  (i.e., an $n$-tuple containing
    $n$ ones).
\end{algorithmic}
}}}
\end{table}

At each communication round agent $i$ executes the algorithm \commrd
displayed in Table~\ref{tab:comm} at the end of the paper.  The following
is an informal description.
\begin{center}
  \noindent{\framebox[.9999\linewidth]{\noindent\parbox{.99\linewidth-2\fboxsep}{
        \textbf{\footnotesize Informal description of \commrd for agent $i$}\\*[0.1cm]
        \footnotesize

        \textbf{Assumes:} $\ind{\status}{i}(s) = 0$ for each
        $s\in\{\ind{\prev}{i}+1,\ind{\prev}{i}+2,\ldots,
        \ind{\n}{i}-1\}\setminus\{\ind{\cu}{i}\} \modulo{n}$.
\begin{algorithmic}[1]
\STATE Broadcast $\ind{\mess}{i}$, consisting of the target indices,
    $\ind{\prev}{i}$, $\ind{\cu}{i}$, and $\ind{\n}{i}$, the UID $i$,
    and the distance to the current target, $\ind{\dist}{i}$.
\FORALL{messages, $\ind{\mess}{k}$, received}
    \STATE Set $\ind{\status}{i}(j)$ to assigned (`0') for each target $j$ from
        $\ind{\prev}{k}+1\modulo{n}$ to $\ind{\n}{k}-1\modulo{n}$ not equal to
        $\ind{\cu}{i}$.
    \IF{$\ind{\prev}{k}=\ind{\n}{k}=\ind{\cu}{k}\neq\ind{\cu}{i}$}
    \STATE Set the status of $\ind{\cu}{k}$ to 0 (because it was missed in the
        previous step).
    \ENDIF
    \IF{$\ind{\cu}{i}=\ind{\cu}{k}$ but agent $i$ is farther from $\ind{\cu}{i}$
        than agent~$k$ (ties broken with UIDs)}
        \STATE Set the status of $\ind{\cu}{i}$ to assigned (`0').
    \ENDIF
    \IF{$\ind{\cu}{i}=\ind{\cu}{k}$ and agent $i$ is closer than agent~$k$}
        \STATE Leave $\ind{\cu}{i}$ unchanged.  However, agent $k$ will set
            $\ind{\cu}{k}$ to a new target.  This target will be at least as far
            along the tour as the farther of $\ind{\n}{i}$ and $\ind{\n}{k}$.  So,
            set the status of $\ind{\n}{i}$ and $\ind{\n}{k}$ to assigned (`0').
    \ENDIF
\ENDFOR
\IF{the status of every target is assigned (`0')}
\STATE Exit \etspa and stop motion. (This can occur only if there are more agents
    than targets and every target is assigned.)
\ELSE
\STATE  Update $\ind{\cu}{i}$ to the next target in the tour with status
    available (`1'), $\ind{\n}{i}$ to the next available target in the tour
    after $\ind{\cu}{i}$, and $\ind{\prev}{i}$ to the first available target
    in the tour before $\ind{\cu}{i}$.
\ENDIF
\end{algorithmic}
}}}
\end{center}

\begin{figure}
\begin{center}
\subfigure[Setup before the conflict over target 7.]
{\includegraphics[width=0.6\linewidth]{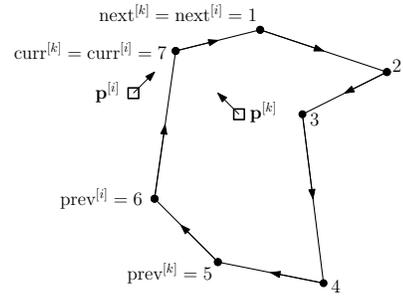}
\label{fig:before_conflict}} \hfill \subfigure[Setup after resolution of
the conflict.] {\includegraphics[width=0.75\linewidth]{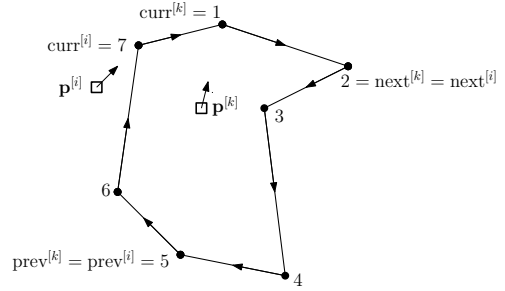}
\label{fig:after_conflict}} \caption{The resolution of a conflict between
agents $i$ and $k$ over
  target $7$.  Since agent $i$ is closer to target $7$ than agent $k$,
  agent $i$ wins the conflict.}
\label{fig:conflict}
\end{center}
\end{figure}

Fig.~\ref{fig:conflict} gives an example of \commrd resolving a conflict
between agents $i$ and $k$, over $\ind{\cu}{i}=\ind{\cu}{k}$. In this
figure, all other agents are omitted.

We are now ready to define the algorithm \etspa for solving the target
assignment problem.
\begin{defn}[\etspa]
  The \etspa algorithm is the triplet consisting of the initialization
  of each agent (see Table \ref{tab:init}), the motion law in
  \eqref{eq:cont_law}, and \commrd (see Table \ref{tab:comm}), which
  is executed at each communication round.
\end{defn}

\subsection{Correctness of \etspa}
\label{sec:correctness} We will now prove the correctness of \etspa.  It
should be noted that this result is valid for any communication graph
which contains the $r$-disk graph as a subgraph. In order to prove
correctness, let us first present some properties of the algorithm.
\begin{lemma}[\etspa properties]
\label{lem:tsp_props} During an execution of \etspa the following
statements hold:
\begin{enumerate}
\item Once target $j\in\I$, is assigned to some agent, the assignment
  may change, but target $j$ remains assigned for all time.
\label{lemit:assigned_invariance}
\item Agent $i$ is assigned to the target $\ind{\cu}{i}$ which
  satisfies $\ind{\status}{i}({\ind{\cu}{i}})=1$.
  \label{lemit:current_1}
\item For agent $i$, $\ind{\status}{i}(j)=0$, for each
  $j\in\{\ind{\prev}{i}+1,\ind{\prev}{i}+2,\ldots,
  \ind{\n}{i}-1\}\setminus\{\ind{\cu}{i}\} \modulo{n}$.
\label{lemit:prev_next_i}
\item For agent $i$, $\ind{\status}{i}(j)=0$ only if target $j$ is
  assigned to some agent $k\neq i$.
\label{lemit:0_only_if}
\item If, for agent $i$, $\ind{\status}{i}(j)=0$ at some time $t_1$,
  then $\ind{\status}{i}(j)=0$ for all $t \geq t_1$.
\label{lemit:status_0}
\item If agent $i$ receives $\ind{\mess}{k}$ during a communication
  round, agent $i$ will set $\ind{\status}{i}(j)=0$ for each
  $j\in\{\ind{\prev}{k}+1, \ldots,
  \ind{\n}{k}-1\}\setminus\{\ind{\cu}{i}\}\modulo{n}$.
\label{lemit:prev_next_k}
\end{enumerate}
\end{lemma}
\begin{proof}
  Statements \ref{lemit:current_1} and \ref{lemit:status_0} and
  \ref{lemit:prev_next_k} follow directly from the initialization and
  the algorithm \commrd.

To see \ref{lemit:assigned_invariance}, consider an agent $i$, who is
assigned to target~$j$.  Agent $i$'s assignment can change only if it
loses a conflict over target $j$.  In every conflict there is a winner and
the winner remains assigned to target $j$.

Statement \ref{lemit:prev_next_i} is initially satisfied since
$\ind{\prev}{i}+1=\ind{\cu}{i}=\ind{\n}{i}-1$ implies that
$\{\ind{\prev}{i}+1,
\ldots,\ind{\n}{i}-1\}\setminus\{\ind{\cu}{i}\}=\emptyset$.  Assume that
statement \ref{lemit:prev_next_i} is satisfied before the execution of
\commrd.  At the end of \commrd, $\ind{\prev}{i}$ is updated to the first
target before $\ind{\cu}{i}$ in the tour with status available (`1').  If
$\ind{\status}{i}(\ind{\cu}{i})=1$ then $\ind{\cu}{i}$ remains unchanged.
If $\ind{\status}{i}(\ind{\cu}{i})=0$ then $\ind{\cu}{i}$ is increased to
the first target with status available (`1').  Finally, $\ind{\n}{i}$ is
set to the first target after $\ind{\cu}{i}$ which is available. Thus, at
the end of \commrd the status of $\ind{\prev}{i}$, $\ind{\cu}{i}$ and
$\ind{\n}{i}$ are available, and $\ind{\status}{i}(j)=0$ for each target
$j\in\{\ind{\prev}{i}+1,\ldots,
\ind{\n}{i}-1\}\setminus\{\ind{\cu}{i}\}\modulo{n}$.

Statement \ref{lemit:0_only_if} is also initially satisfied since
$\ind{\status}{i}=\mathbf{1}_n$ for each $i\in\I$.  Assume Statement
\ref{lemit:0_only_if} is satisfied before the execution of \commrd and
that during this communication round agent $i$ changes the status of a
target $j$ to assigned (`0').  We will show that Statement
\ref{lemit:0_only_if} is still satisfied upon completion of the execution
of \commrd.  In order for $\ind{\status}{i}(j)$ to be changed, agent $i$
must have received a message, $\ind{\mess}{k}$, for which one of the
following cases is satisfied: (1) Target $j \neq \ind{\cu}{i}$ lies
between $\ind{\prev}{k}$ and $\ind{\n}{k}$ on the tour; (2) There is a
conflict between agents $i$ and $k$ over target $j$ which agent $i$ loses;
or, (3) There is a conflict between agents $i$ and $k$ which agent $i$
wins and $\ind{\n}{i}=j$ or $\ind{\n}{k}=j$.

In Case (1) either $\ind{\status}{k}(j)=0$ or $\ind{\cu}{k}=j$, and thus
target $j$ is assigned.  In Case (2) agent $k$ won the conflict implying
$\ind{\cu}{k}=j$ entering the communication round.  Thus after the
communication round, $\ind{\cu}{i}\neq j$ and target $j$ is assigned to
another agent.  In Case (3), $\ind{\cu}{i}=\ind{\cu}{k}\neq j$, and agent
$k$ loses the conflict. In this case, agent $k$ will change $\ind{\cu}{k}$
to the next available target on its tour.  All targets from
$\ind{\prev}{k}+1$ to $\ind{\n}{k}-1$ have been assigned.  Also, during
the communication round, agent $k$ will receive $\ind{\mess}{i}$ and
determine that all targets from $\ind{\prev}{i}+1$ to $\ind{\n}{i}-1$ are
assigned. Thus, the next available target is at least as far along the
tour as the farther of $\ind{\n}{i}$ and $\ind{\n}{k}$. Thus, after the
communication round, both $\ind{\n}{i}$ and $\ind{\n}{k}$ are assigned.
\end{proof}

With these properties we are now ready to present the main result of this
section.
\begin{theorem}[Correctness of \etspa]
\label{thm:strategy_works} For any fixed $n\in\N$, \etspa solves the
target assignment problem.
\end{theorem}
\begin{proof}
  Assume by way of contradiction that at some time $t_1 \geq 0$ there
  are $J\in \{1,\ldots,n-1\}$ targets unassigned, and for all time $t
  \geq t_1$, $J$ targets remain unassigned.  By Lemma
  \ref{lem:tsp_props}~\ref{lemit:assigned_invariance} the $n-J$
  assigned targets remain assigned for all time, and thus it must be
  the same $J$ targets which remain unassigned for all $t\geq t_1$.
  Let $\J$ denote the index set of the $J$ unassigned targets.  From
  our assumption, and by Lemma
  \ref{lem:tsp_props}~\ref{lemit:0_only_if}, for every $t\geq t_1$ and
  for every $i\in\I$, $\ind{\status}{i}(j)=1$ for each $j\in\J$.  Now,
  among the $n-J$ assigned targets, there is at least one target to
  which two or more agents are assigned.  Consider one such target,
  call it $j_1$, and consider an agent $i_1$ with
  $\ind{\cu}{i_1}=j_1$.  By Lemma \ref{lem:com_range}, agent $i_1$
  will enter a conflict over $j_1$ in finite time.  Let us follow the
  loser of this conflict. The losing agent, call it $i_2$, will set
  $\ind{\status}{i_2}(j_1)=0$, and will move to the next target in the
  tour it believes may be available, call it $j_2$.  Now, we know
  $j_2$ is not in $\J$, for if it were $J-1$ targets would be
  unassigned contradicting our assumption.  Moreover, by Lemma
  \ref{lem:tsp_props}~\ref{lemit:current_1}, $j_2\neq j_1$.  Thus,
  agent $i_2$ will enter a conflict over $j_2$ in finite time.  After
  this conflict, the losing agent, call it $i_3$, will set
  $\ind{\status}{i_3}(j_2)=0$ (because it lost the conflict), and from
  Lemma \ref{lem:tsp_props}~\ref{lemit:prev_next_k},
  $\ind{\status}{i_3}(j_1)=0$.  Again, agent $i_3$'s next target,
  $j_3$ must not be in $\J$, for if it were we would have a
  contradiction.  Thus, repeating this argument $n-J$ times we have
  that agent $i_{n-J}$ loses a conflict over $j_{n-J}$.  After this
  conflict, we have $\ind{\status}{i_{n-J}}(j_k)=0$ for each
  $k\in\{1,\ldots,n-J\}$, where $j_{k_1}=j_{k_2}$ if and only if
  $k_1=k_2$.  In other words, agent $i_{n-J}$ knows that all $n-J$
  assigned targets have indeed been assigned.  Also, by our initial
  assumption, $\ind{\status}{i_{n-J}}(j)=1$ for each $j\in\J$.  By
  Lemma \ref{lem:tsp_props}~\ref{lemit:current_1}, agent $i_{n-J}$'s
  new current target must have status available (`1').  Therefore, it
  must be that agent $i_{n-J}$ will set $\ind{\cu}{i_{n-J}}$ to a
  target in $\J$.  Thus, after a finite amount of time, $J-1$ targets
  are unassigned, a contradiction.
\end{proof}

The following remark displays that the \etspa algorithm does not solve the
target assignment under the consistent knowledge assumption.
\begin{remark}[Consistent knowledge: cont'd]
\label{rem:not_consist_knowledge} Consider as in
Remark~\ref{rem:consist_knowledge} the consistent knowledge assumption for
each agent's target set.  Specifically, consider two agents, 1 and 2, with
initial target sets $\ind{\tarset}{1}(0)=\{\q_2\}$,
$\ind{\tarset}{2}(0)=\{\q_1,\q_2\}$, and any initial positions such that
$\ind{\p}{1}(0)=\q_2$, We will have $\ind{\cu}{i}=\ind{\cu}{j}=2$.
However, agent $2$ will win the conflict over target 2.  Thus, agent 1
will set $\ind{\status}{1}(2)=0$, and a complete assignment will not be
possible.  \oprocend
\end{remark}

\subsection{Time complexity for \etspa}
\label{sec:time_complex} In this section we will give an upper bound on
the time complexity for \etspa. We will show that when $\ell(n)\geq
(1+\epsilon)rn^{1/d}$, for some $\epsilon\in\R_{>0}$, \etspa is
asymptotically optimal among algorithms in the \algclass class.  Before
doing this, let us first comment on the lower bound when the environment
grows at a slower rate.

In what follows we show that if an agent arrives and remains at its
assigned target for sufficiently long time, then it stays there for all
subsequent times.
\begin{lemma}
\label{lem:stop} Consider $n$ agents executing \etspa with communication
range $r>0$ and assume the time delay between communication rounds,
$t_{max}$, satisfies $t_{max} <r/v$.  If there exists a time $t_1$ and an
agent $i$ such that $\ind{\p}{i}(t)=\ind{\cu}{i}$ for all
$t\in[t_1,t_1+t_{max}]$, then $\ind{\p}{i}(t)=\ind{\cu}{i}$ for all
$t>t_1+t_{max}$.
\end{lemma}
\begin{proof}
  Consider agent $i$, who has been at target $\ind{\cu}{i}$ during the
  entire time interval $[t_1,t_1+t_{max}]$.  By the definition of
  $t_{max}$ there was a communication round at some time $t_2\in
  {[t_1,t_1+t_{max}[}$.  Agent $i$ must have won any conflicts it
  entered during this communication round since we have assumed that
  $\ind{\p}{i}(t_1+t_{max}) = \ind{\cu}{i}$.  Thus every agent $k$
  within distance $r$ of $\ind{\cu}{i}$ will have set
  $\ind{\status}{k}(\ind{\cu}{i})=0$.  After the communication round
  at $t_2$, every agent $k$ with $\ind{\cu}{k}=\ind{\cu}{i}$ must be a
  distance greater than $r$ from $\ind{\cu}{i}$.  Since $t_{max}
  <r/v$, any agent $k$ that enters a conflict with agent $i$ at time
  $t >t_2$, will be at a distance $\ind{\dist}{k}\in{]0,r[}$ from
  $\ind{\cu}{i}$.  Agent $k$ will lose the conflict since
  $\ind{\dist}{k} >0=\ind{\dist}{i}$.  Thus, agent $i$ will remain at
  $\ind{\cu}{i}$ for all $t>t_1+t_{max}$.
\end{proof}
With this lemma we are now able to provide an upper bound on the time
complexity of our scheme.
\begin{theorem}[Time complexity for \etspa]
\label{thm:tight_bound} Consider an environment $\env=[0,\ell(n)]^d$,
$d\geq 1$. If $t_{max} <r/v$, then \etspa solves the target assignment
problem with time complexity in $O(n^{(d-1)/d}\ell(n) +n)$.  If, in
addition, $\ell(n)\geq(1+\epsilon)r n^{1/d}$, where $\epsilon
\in\mathbb{R}_{>0}$, the time complexity is in
$\Theta(n^{(d-1)/d}\ell(n))$, and \etspa is asymptotically optimal among
algorithms in the \algclass class.
\end{theorem}
\begin{proof}
  Consider any initial agent positions,
  $\ind{p}{1}(0),\ldots,\ind{p}{n}(0)$, and any $n$-tuple of target
  positions, $\tar$.  In the worst-case, some agent must travel around
  its entire ETSP tour, losing a conflict at each of the first $n-1$
  targets in the tour.  By Lemma \ref{lem:stop}, this agent can spend
  no more than $t_{max}$ time units at each of the $n-1$ targets,
  before losing a conflict.  Since each agent's tour is a constant
  factor approximation of the optimal, the tour length is
  $O(n^{(d-1)/d}\ell(n))$ (see Theorem \ref{thm:tsp_grow}).  The agent
  will not follow the ETSP tour exactly because it will enter a
  conflict over each of the $n-1$ targets before actually reaching the
  target.  However, the resulting path is no longer than the ETSP tour
  (since the agent could just follow the ETSP tour exactly if that
  happened to be the shortest path).  Hence, the time complexity is
  $O(n^{(d-1)/d}\ell(n) +t_{max}(n-1))\in O(n^{(d-1)/d}\ell(n) +n)$.
  If $\ell(n)=(2+\epsilon)r n^{1/d}$, with $\epsilon
  \in\mathbb{R}_{>0}$, we can combine this with Theorem
  \ref{thm:comp_lower} to get a time complexity in
  $\Theta(n^{(d-1)/d}\ell(n))$.
\end{proof}
Notice that when $\ell(n)$ satisfies the bound in Theorem
\ref{thm:tight_bound}, and $\ell(n)\in O(n^{1/d})$, the time complexity is
in $O(n)$.

We have given complexity bounds for the case when $r$ and $v$ are fixed
constants, and $\ell(n)$ grows with $n$.  We allow the environment
$\env(n)$ to grow with $n$ so that, as more agents are involved in the
task, their workspace is larger.  An equivalent setup would be to consider
$\ell$ to be fixed, and allow $r$ and $v$ to vary inversely with the $n$.
That is, we can introduce a set of parameters, $\tilde \ell=1$, and
$\tilde r(n)$ and $\tilde v(n)$ such that the time complexity will be the
same as for the parameters $r$, $v$, $\ell(n)$.

\begin{corollary}[Scaling radius and speed]
\label{cor:unit_cube_env} Consider $n$ agents in the environment
$\env=[0,1]^d$, with speed $\tilde v(n):=v/\ell(n)$, and communication
radius $\tilde r(n):=r/\ell(n)$, where $\ell(n)\geq (1+\epsilon)r
n^{1/d}$, and $\epsilon \in\mathbb{R}_{>0}$.  Then \etspa solves the
target assignment problem with time complexity in
$\Theta(n^{(d-1)/d}\ell(n))$.
\end{corollary}

Scaling the communication radius $r$ inversely with the number of agents
arises in the study of wireless networks \cite{PG-PRK:00}.  In wireless
applications there are interference and media access problems between
agents in the network.  Since the agents are in a compact environment, the
only way to limit this interference is to scale the communication radius
inversely with the number of agents.  Scaling the agent speed inversely
with $n$ appears in the study of the vehicle routing problem in
\cite{VS-MS-EF-PV:05a}.  The inverse scaling is required to avoid
collisions in the presence of traffic congestion.

\subsection{Communication and computation complexity}
\label{sec:comm_comp_complexity} In our notion of time complexity we have
emphasized the complexity due to the motion of the agents. Here we will
briefly classify the complexity of computation and communication for
\etspa. (i) \emph{Initialization:} As reviewed in Section
\ref{sec:TSP_review}, we can compute a constant factor approximation ETSP
tour in time $O(n^2)$.  This is the most expensive computation and thus
the complexity of initialization is in $O(n^2)$.  (ii) \emph{Communication
  complexity per round:} At each round agent $i$ broadcast a message
of length $O(\log{n})$, $\ind{\mess}{i}$, and we consider this to be one
unit of communication.  In the worst-case, each agent receives $n$
messages, and so, the worst-case communication complexity is in $O(n)$
\cite{SM-FB-JC-EF:05mn-tmp}.  (iii) \emph{Computation complexity per
  round:} For each message received, agent $i$ sets
$\ind{\status}{i}(s)=0$ for $s$ from $\ind{\prev}{k}+1$ to
$\ind{\n}{k}-1$.  In the worst-case, this operation is $O(n)$ and must be
performed for $n$ messages.  This is the dominant computation in \commrd
and thus the worst-case computation complexity in each round is $O(n^2)$.

It should be noted that in the case when the communication graph is not
even connected (let alone complete as is required to achieve these
worst-case bounds), the complexity will be considerably lower.
\subsection{Simulations}
\label{sec:simulations} 
\begin{figure}
\begin{center}
\subfigure[Initial agent and target positions.]
{\includegraphics[width=0.47\linewidth]{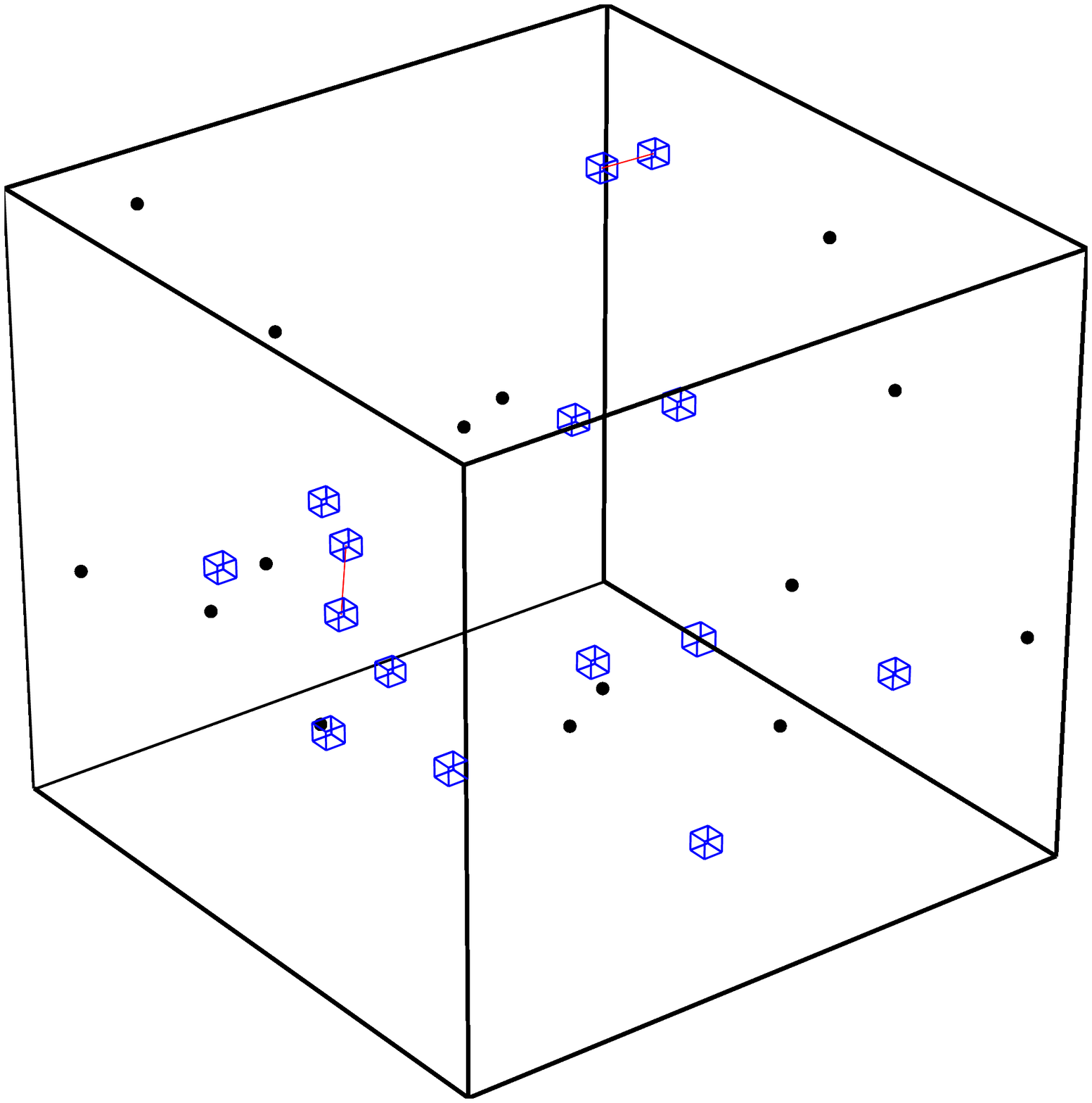} \label{fig:initial}}
\hfill \subfigure[Positions after 30 time units.]
{\includegraphics[width=0.47\linewidth]{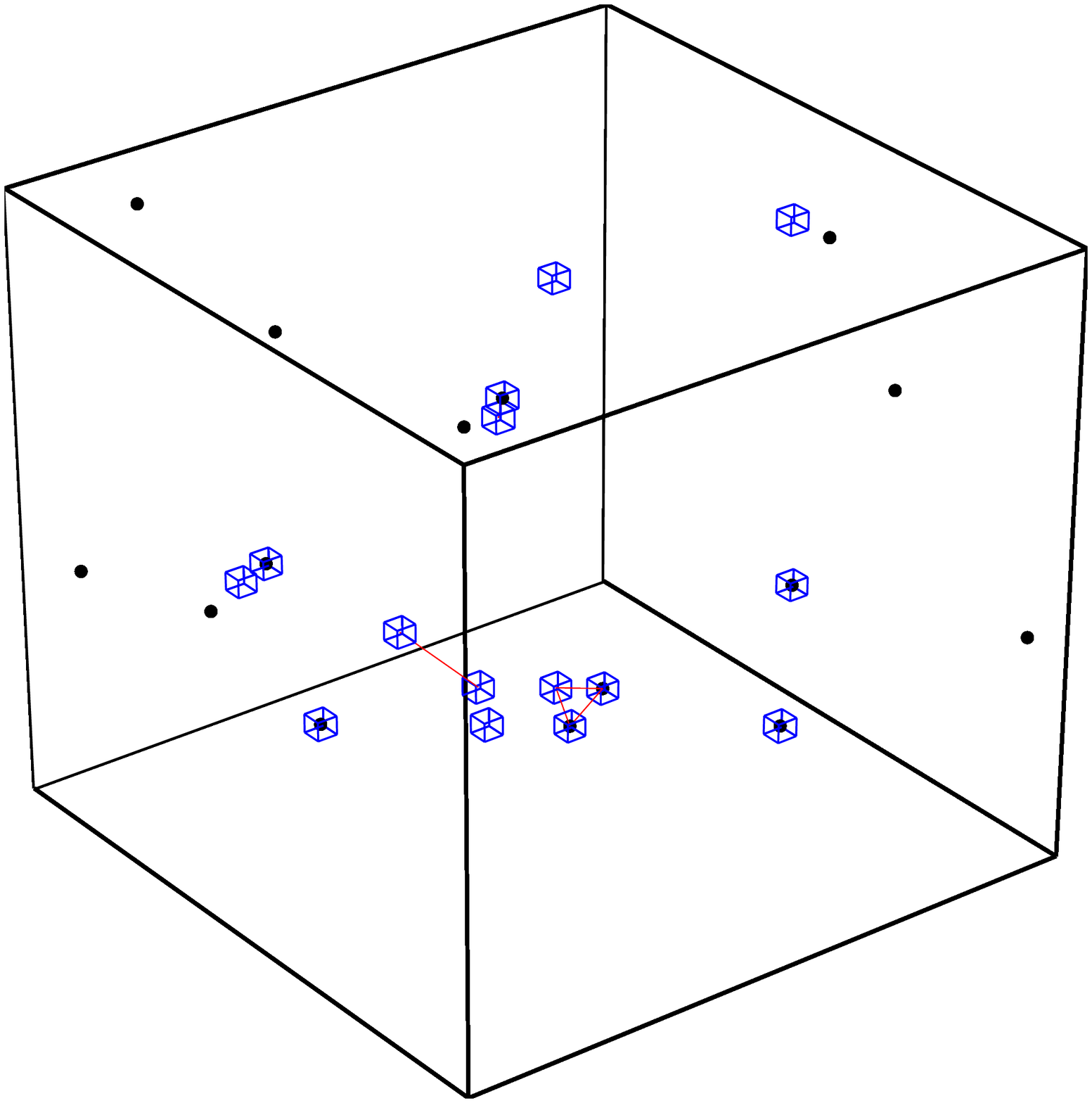}
\label{fig:frame_31}} \subfigure[Positions after 90 time units.]
{\includegraphics[width=0.47\linewidth]{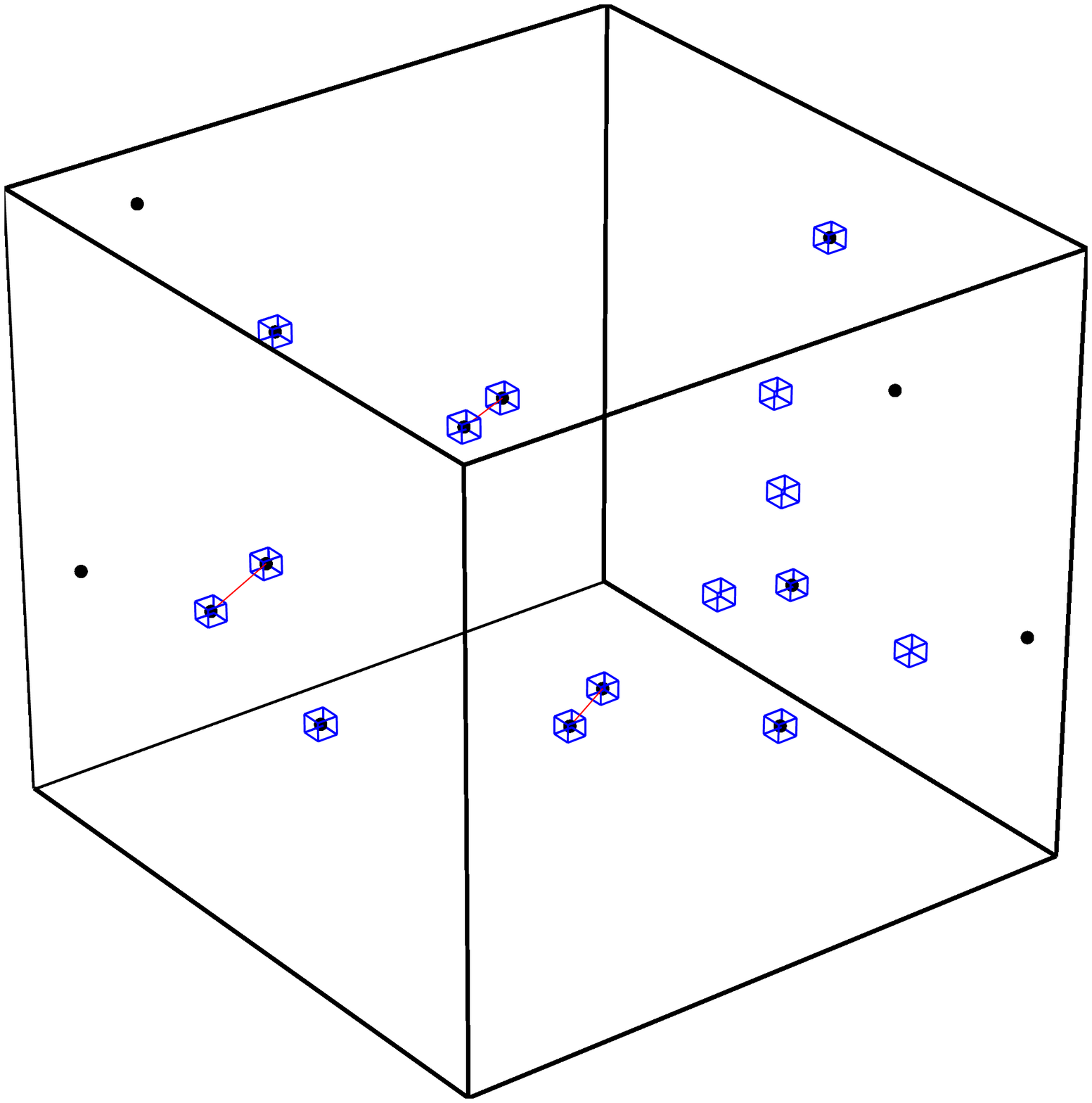}
\label{fig:frame_91}} \hfill \subfigure[Complete target assignment.]
{\includegraphics[width=0.47\linewidth]{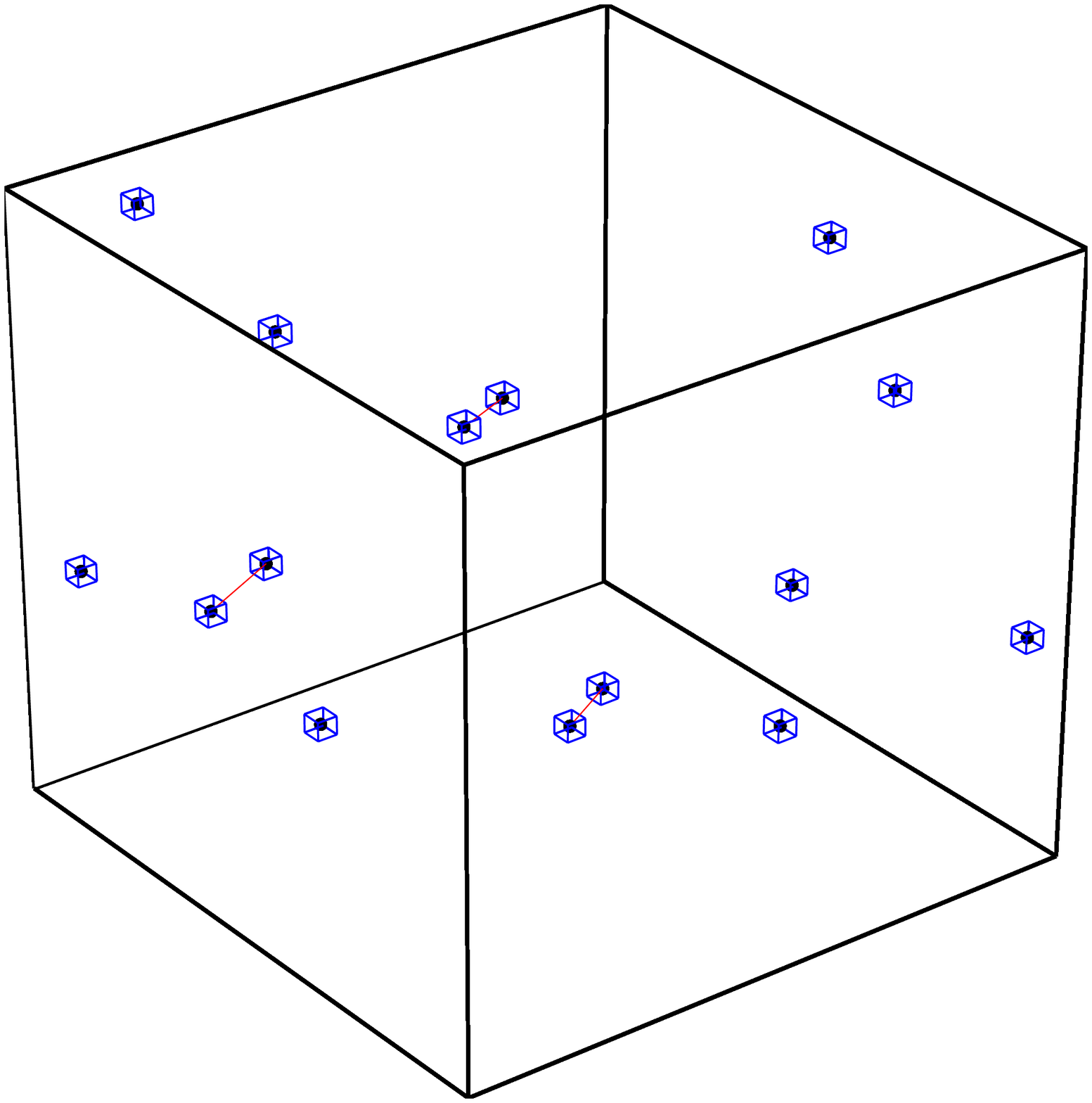} \label{fig:final}}
\caption{Simulation for 15 agents, with $v=1$ and $r=15$.  in an
  environment $[0,100]^3$. The targets are spheres and the agents are
  cubes.  An edge is drawn between two agents when they are within
  communication range.}
\label{fig:simu3D}
\end{center}
\end{figure}

We have simulated \etspa in $\R^2$ and $\R^3$.  To
compute the ETSP tour we have used the \texttt{concorde} TSP
solver.\footnote{The
  \texttt{concorde} TSP solver is available for research use at
  \texttt{http://www.tsp.gatech.edu/concorde/index.html}} A
representative simulation for 15 agents in $[0,100]^3\subset\R^3$ with
$r=15$ and $v=1$ is shown in Fig. ~\ref{fig:simu3D}.  The initial
configuration shown in Fig.~\ref{fig:initial} consists of uniformly
randomly generated target and agent positions.

\subsection{The case of $n$ agents and $m$ targets}
\label{sec:n_neq_m} It should be noted that the \etspa algorithm works
without any modification when there are $n$ agents and $m$ targets.  If
$m\geq n$, at completion, $n$ targets are assigned and $m-n$ targets are
not. When, $m <n$, at completion, all $m$ targets are assigned, and the
$n-m$ unassigned agents come to a stop after losing a conflict at each of
the $m$ targets. The complexity bounds are changed as follows.

The lower bound on the \algclass class in Theorem \ref{thm:comp_lower},
holds when $m \geq n$, and $\ell(n) \geq (1+\epsilon)r m^{1/d}$ (notice
the $m$ instead of $n$).  The bound becomes $\Omega(\ell(n) m^{-1/d}n)$.
If $m= Cn$ where $C \in\mathbb{R}_{\geq 1}$, (i.e., $m \geq n$ but they
grow at the same rate), then the bound becomes $\Omega(\ell(n)
n^{(d-1)/d})$.

The upper bound on \etspa holds for any $n$ and $m$, and becomes
$O(\ell(n) N^{(d-1)/d} )$, where $N:=\min\{n,m\}$.  So our final result
would be that if $m= Cn$ where $C \in\mathbb{R}_{\geq 1}$ and when
$\ell(n) \geq (1+\epsilon)r m^{1/d}$, then \etspa solves the target
assignment problem in $\Theta ( \ell(n) n^{(d-1)/d} )$.  That is, among
all algorithms in the \algclass class, \etspa is asymptotically optimal.

\section{Conclusions}
\label{sec:concl} We have developed the \etspa algorithm for solving the
full knowledge target assignment problem.  We derived worst-case
asymptotic bounds on the time complexity, and we showed that among a
certain class of algorithms, \etspa is asymptotically optimal.  There are
many possible extensions of this work.  We have not given a lower bound on
the time complexity of \etspa when $\ell(n)\leq \ell_{crit}$.  Also, the
problem is unsolved under the more general consistent knowledge
assumption. We would like to extend the \etspa algorithm to agents with
nonholonomic motion constraints. Also, it would be interesting to consider
a sensor based version of this problem, where agents acquire target
positions through local sensing.  Finally, to derive asymptotic time
bounds, we made some assumptions on the communication structure at each
communication round. An interesting avenue for future study would be to
more accurately address the communication issues in robotic networks.
\begin{table}[ht]
\caption{\label{tab:comm} Communication Round (\commrd) for agent $i$.}
\begin{center}
  \noindent{\framebox[.9999\linewidth]{\noindent\parbox{.9999\linewidth-2\fboxsep}
      {%
        \noindent\begin{tabular}{ll} \textbf{Name:} &
          \parbox[t]{.75\linewidth}{\commrd}
          \\
          \textbf{Goal:} & \parbox[t]{.75\linewidth}{Obtain
            information on assigned targets.}
          \\
          \textbf{Assumes:} & \parbox[t]{.75\linewidth}{(i) Knowledge
            of the $n$-tuple $\tar$, and a method for computing a
            constant factor TSP tour of the $n$ targets, $\tour$.
            (ii) A communication range $r > 0$.}  \vspace{.4em}
          \\
          \hline
        \end{tabular}\\[.5ex]
        \begin{algorithmic}[1]
          \STATE Compute
          $\ind{\dist}{i}:=\|\ind{\p}{i}-\ind{\tar}{i}_{\ind{\cu}{i}}\|$.
          \STATE Broadcast
          $\ind{\mess}{i}:=(\ind{\prev}{i},\ind{\cu}{i},\ind{\n}{i},i,\ind{\dist}{i})$.
          \STATE Receive $\ind{\mess}{k}$, from each $k\neq i$
          satisfying $\|\ind{\p}{i}-\ind{\p}{k}\|\leq r$.
          \FORALL{$\ind{\mess}{k}$ received} \FOR{$s =
            \ind{\prev}{k}+1$ to $\ind{\n}{k}-1 \modulo{n}$}
          \IF{$s\neq \ind{\cu}{i}$} \STATE Set $\ind{\status}{i}(s)
          :=0$
          \ENDIF
          \ENDFOR
          \IF{$\ind{\prev}{k}=\ind{\n}{k}=\ind{\cu}{k}\neq\ind{\cu}{i}$}
          \STATE Set $\ind{\status}{i}(\ind{\cu}{k}) := 0$
          \ENDIF
          \IF{$\ind{\cu}{i}=\ind{\cu}{k}$}
          \IF{($\ind{\dist}{i} >
            \ind{\dist}{k}$) OR ($\ind{\dist}{i} = \ind{\dist}{k}$ AND
            $i < k$)}
            \STATE Set $\ind{\status}{i}({\ind{\cu}{i}}):=0$.
          \ELSE
          \IF{$\ind{\n}{i} \neq \ind{\cu}{i}$}
          \STATE Set $\ind{\status}{i}({\ind{\n}{i}}):=0$.
          \ENDIF
          \IF{${\ind{\n}{k}} \neq {\ind{\cu}{i}}$}
          \STATE Set $\ind{\status}{i}({\ind{\n}{k}}):=0$.
          \ENDIF
    \ENDIF
\ENDIF
\ENDFOR
\IF{$\ind{\status}{i}(j)=0$ for every target $j$}
\STATE Exit \etspa and stop motion.
\ELSE
\WHILE{$\ind{\status}{i}({\ind{\cu}{i}})$=0}
    \STATE $\ind{\cu}{i}:=\ind{\cu}{i}+1\modulo{n}$.
\ENDWHILE
\STATE Set $\ind{\n}{i}:=\ind{\cu}{i}+1 \modulo{n}$.
\WHILE{$\ind{\status}{i}({\ind{\n}{i}})$=0}
    \STATE $\ind{\n}{i}:=\ind{\n}{i}+1\modulo{n}$.
\ENDWHILE
\WHILE{$\ind{\status}{i}({\ind{\prev}{i}})$=0}
    \STATE $\ind{\prev}{i}:=\ind{\prev}{i}-1\modulo{n}$.
\ENDWHILE
\ENDIF
\end{algorithmic}
}}}
\end{center}
\vspace{-3ex}\end{table}


\end{document}